\RequirePackage{amssymb}
\documentclass[sigconf,natbib=true]{acmart}
\usepackage{preprint_header}

\AtBeginDocument{%
  \providecommand\BibTeX{{%
    \normalfont B\kern-0.5em{\scshape i\kern-0.25em b}\kern-0.8em\TeX}}}

\usepackage{setspace}
\usepackage[most]{tcolorbox}
\setcopyright{acmlicensed}
\copyrightyear{2024}
\acmYear{2024}
\acmDOI{XXXXXXX.XXXXXXX}

\acmConference[SIGIR 2024]{The 47th International ACM SIGIR Conference on Research and Development in Information Retrieval}{July 14--18, 2024}{Washington D.C., USA}
\acmISBN{978-1-4503-XXXX-X/18/06}

\begin{document}

\title{Can LLMs Master Math? Investigating Large Language Models on Math Stack Exchange}


\author{Ankit Satpute$^{\dag}$$^{\S}$, Noah Gießing$^{\dag}$, Andr\'{e} Greiner-Petter$^{\S}$, Moritz Schubotz$^{\dag}$, Olaf Teschke$^{\dag}$, Akiko Aizawa$^{\pounds}$, Bela Gipp$^{\S}$}
\email{Ankit.Satpute@fiz-karlsruhe.de}
\affiliation{
  \institution{$^{\dag}$FIZ Karlsruhe Berlin, $^{\S}$University of Göttingen Germany, $^{\pounds}$NII Japan}
\country{}
}

\renewcommand{\shortauthors}{Satpute, et al.}

\begin{abstract}
Large Language Models (LLMs) have demonstrated exceptional capabilities in various natural language tasks, often achieving performances that surpass those of humans. 
Despite these advancements, the domain of mathematics presents a distinctive challenge, primarily due to its specialized structure and the precision it demands. 
In this study, we adopted a two-step approach for investigating the proficiency of LLMs in answering mathematical questions.
First, we employ the most effective LLMs, as identified by their performance on math question-answer benchmarks, to generate answers to 78 questions from the Math Stack Exchange (MSE).
Second, a case analysis is conducted on the LLM that showed the highest performance, focusing on the quality and accuracy of its answers through manual evaluation.
We found that GPT-4 performs best (nDCG of 0.48 and P@10 of 0.37) amongst existing LLMs fine-tuned for answering mathematics questions and outperforms the current best approach on ArqMATH3 Task1, considering P@10.
Our Case analysis indicates that while the GPT-4 can generate relevant responses in certain instances, it does not consistently answer all questions accurately.
This paper explores the current limitations of LLMs in navigating complex mathematical problem-solving. 
Through case analysis, we shed light on the gaps in LLM capabilities within mathematics, thereby setting the stage for future research and advancements in AI-driven mathematical reasoning.
We make our code and findings publicly available for research: \url{https://github.com/gipplab/LLM-Investig-MathStackExchange}
\end{abstract}



\maketitle
\thispagestyle{preprintbox}

\section{Introduction}


Large Language Models (LLMs) have been very popular because of their ability to solve natural language tasks, sometimes with human-like accuracy~\cite{23LLMsuccess,23SIGIRLLMsuccess,2023LLMsuccess_n}. 
Pre-training and fine-tuning the models have led to a performance-winning approach in tasks such as translation, writing code, passing professional exams, etc~\cite{23LLMPreandFIneperf}.
The LLMs are used in academia and scientific research for knowledge extraction, ideas, processing, comparison, and across multiple disciplines~\cite{23LLMinAcademia}.
LLms have also been successful in Question-Answering tasks (QA) in which they provide human-like answers to questions in natural language~\cite{21LMforQA}.
Evaluation of LLMs on QA has been especially useful in detecting how accurate LLMs are in generating answers and finding cases in which it could halllucinate~\cite{24LLMunderstanding,23evaluatingLLMQa}.

Given that mathematical content is prevalent in science, technology, engineering, and mathematics (STEM), it is imperative to evaluate the capacity of LLMs to engage with mathematical language.
The challenges of LLMs with mathematics are manifold; mathematics, with its rigorous logic and abstract concepts, is communicated through a specialized language that mixes symbols and syntax in complex arrangements~\cite{Satpute2024}.
Unlike natural language, mathematical expressions rely on unmentioned rules and assumptions, demanding explicit knowledge and a level of precision.
Mathematical reasoning still poses a sizable challenge to even the most significant language models~\cite{24LLMMathCompet,24MathAnswersLLM}.
There exists an evaluation of LLMs in answering math questions in which the expected answer is straightforward numerical value~\cite{21MATHDataset}, and an LLM-based solution~\cite{23Tora} is the leading method on the performance scoreboard.
In contrast, answers for proof or conceptual questions, which form a large part of the MSE, are not verifiable without human aid yet.
However, the evaluation of models in generating answers for unorganized open questions has not been performed.

In this paper, we investigate using LLMs to answer open-ended questions (questions that cannot be answered with a yes or no response) from mathematics. 
We utilized questions from Math Stack Exchange (MSE), a platform dedicated to solving a wide array of mathematical questions posed by users, offering an ideal testing ground.
We evaluate LLMs to find the most relevant answer.
MSE encompasses questions that range from elementary level to advanced mathematics
MSE questions require correctly applying mathematical principles and articulating complex reasoning clearly and understandably.
By focusing on open-ended questions from MSE, we not only underscore the importance of these questions in fostering a deeper understanding of mathematical concepts but also provide a benchmark against which the progress of LLMs in mathematical reasoning can be measured.
In this work, we set a base to explore the path forward for LLMs by evaluating them on generated answers and highlighting the challenges using a case study, thus bridging the gap between natural and mathematical language comprehension for LLMs.
We make our code and findings publicly available for \textbf{research}\footnote{\url{https://github.com/gipplab/LLM-Investig-MathStackExchange}\label{repoCodeData}}.


\section{Related work}

Research on Large Language Models (LLMs) has extended into mathematical problem solving, albeit primarily focused on pre-university mathematics. 
Several datasets featuring high-school-level math questions and answers exist, such as those introduced by Mao et al.~\cite{mao2024champ}, Urrutia et al.~\cite{urrutia2023whos}, and Deb et al.~\cite{deb2023blank}.
Hendrycks et al.~\cite{21MATHDataset} introduced the MATH dataset with 12,500 challenging competition mathematics problems.
Cobbe et al.~\cite{cobbe2021training} developed the GSM8K dataset containing a linguistically diverse collection of high school math problems.
MathQA dataset by Amini et al.~\cite{amini2019mathqa} contains 37k English multiple-choice math word problems covering multiple math domain categories by modeling operation programs corresponding to word problems in the AQuA dataset~\cite{2017AQuA}.
The availability of datasets is hindered as primarily physical records are used for writing question answers instead of machine-processable and computer-aided resources~\cite{beck2020transforming,2023teimma}.

Specialized LLMs have been developed for particular mathematical fields.
For instance, Trinh et al.~\cite{24AlphaGeometry} released AlphaGeometry, a theorem prover for Euclidean plane geometry, which solved 25 out of 30 latest Olympiad-level problems within the time limit.
To mitigate the issue of LLMs generating inaccurate information, Romera-Paredes\cite{24FunSearch} developed FunSearch, which combines a pre-trained LLM with a systematic evaluator to outline verified problem-solving approaches. 
Moreover, LLMs equipped with integrated verification codes have demonstrated remarkable success on the MATH and GSM8K datasets, achieving accuracies near 100\% in some cases~\cite{zhou2023solving}.
The ToRA series~\cite{23Tora}, based on LLaMA-2~\cite{touvron2023llama} and CodeLLaMA~\cite{rozière2024code}, represents the forefront of open-source models on the MATH dataset.
ToRA models are based on LLaMA-2~\cite{touvron2023llama} and CodeLLaMA~\cite{rozière2024code} and were finetuned on a dataset of reasoning paths produced by GPT-4~\cite{openai2023gpt4} for the MATH and GSM-8 datasets. 
These models are fine-tuned on reasoning paths generated by GPT-4 for the MATH and GSM-8 datasets and self-verify their solutions by generating and executing code. 
However, leveraging executable code for Multiple-Selection Questions (MSE) potentially poses challenges due to the complexity of verifying reasoning steps and has not been evaluated so far.
The MAmmoTH~\cite{23mammoth} family of models is trained on a dataset of chain-of-thought and program-of-thought rationales produced by GPT-4, but draws from a more diverse distribution of questions that were asked.
The Llemma~\cite{23lemma} family of models have instead been finetuned on a general corpus of math-related documents and is intended to be a generalist model, leaving open the option of finetuning to specific tasks.

\section{Dataset}

Manual verification of answers to questions from the Mathematics Stack Exchange (MSE) is impractical due to the interdisciplinary nature of the questions and the expertise required to assess answer correctness. 
The ArqMATH competition dataset~\cite{23ArqMATHComp} offers a collection of MSE question-answer pairs. 
Task 1 from the competition's third edition focused on retrieving relevant answers from MSE for 78 undergraduate-level mathematics questions. 
The relevancy of the top-ranked answers, as determined by competing systems, was assessed by student evaluators, averaging 450 evaluated answers per topic. 
While automated evaluation is feasible by excluding answers without evaluation scores, student assessments have shown some inaccuracies~\cite{scharpf2020arqmath}.
\vspace{-2mm}

\section{Methodology}
For our evaluation, we select 6 LLMs.
ToRA~\cite{23Tora}, LLeMa~\cite{23lemma}, GPT-4~\cite{openai2023gpt4}, and MAmmoTH~\cite{23mammoth} give their performance of existing MathQA datasets.
We also consider MABOWDOR~\cite{23mabowdor}, the best-performing approach for ArqMATH Task-1.
Last, we consider Mistral 7B~\cite{jiang2023mistral}, general purpose LLM that has shown sub-level performance with LLaMa, GPT-4 in prompt-based tasks.

We explore two scenarios for our experiments by employing LLMs to answer MSE questions. 
We perform a two-step procedure in the first scenario (Answer generation).
First, we gave 78 questions and generated answers using selected LLMs except MABOWDOR.
MABOWDOR uses BERT-based Desne Passage Retrieval, which can only generate embeddings.
Second, we indexed the answer as embedding and all the other answers from ArqMATH and found the answer most similar to the generated answer from ArqMATH.
In the second scenario (Question-Answer comparison), we use selected LLMs to generate embeddings of all potential answers from ArqMATH and generate embeddings of 78 questions as well.
Ultimately, we find out which answer is most similar to the question.

\section{Evaluation}

This section presents the evaluation of two scenarios using metrics such as Mean Average Precision (mAP'), Precision@10 (P@10), normalized Discounted Cumulative Gain (nDCG'), and Binary Preference (BPref), with scores derived exclusively from evaluated ArqMATH dataset answers.
\vspace{-2mm}

\subsection{Answer generation}

In this phase, answers were generated using the six selected models. These models' outputs served as queries for retrieval within the ArqMATH answer pool. 
For retrieval, we used DPR vector embeddings~\cite{23mabowdor} and cosine similarity.
Table~\ref{satable} displays the comparative performance across all models. 
Additionally, we executed a DPR run for comparison, akin to the method utilized by MABOWDOR.
The outcome reveals that models fine-tuned on mathematical tasks underperformed relative to the DPR benchmark. 
Among the three 7b models (Tora, LLeMa, and MAmmoTH), variations in their performance align with their results on the MATH dataset. 
Notably, increasing the model size of the top performer did not yield better results. 
The Mistral~\cite{jiang2023mistral} model, despite being the lowest scorer on the MATH dataset among the models considered, delivered performance comparable to that of Tora-7b. 
This suggests that models excelling on the MATH dataset might be overfitted to that particular task. 
Conversely, GPT-4 generated answers exhibited increased effectiveness over the DPR baseline, outperforming the current best approach on ArqMATH3 Task1, i.e., MABOWDOR~\cite{23mabowdor} considering P@10.

\begin{table}
\begin{tabular}{lcccc}
\hline
Model       & nDCG' & mAP'  & p@10   & BPref  \\
\hline
DPR (Cocomae)    & 0.464 & 0.191 & 0.324 & 0.192 \\
Tora-7b-Code & 0.400 & 0.159 & 0.279 & 0.170 \\
Tora-13b-Code & 0.355 & 0.140 & 0.266 & 0.154 \\
LLema-7b     & 0.194 & 0.069 &  0.157 & 0.092 \\
MAmmoTH-7b-Coder& 0.369 & 0.145 & 0.253 & 0.166 \\
Mistral-7b-Instruct & 0.396 & 0.155 & 0.267 & 0.166 \\
GPT-4*  (right trunc.)     & \textbf{0.486} & \textbf{0.219} & \textbf{0.374} & \textbf{0.225} \\
GPT-4§ (left trunc.) & 0.473 & 0.210 & 0.367 & 0.215 \\
\hline
\end{tabular}
\caption{Performance of Large Language Models in generating accurate answers.(*: Answer truncated to initial 512 tokens, §:answer truncated to last 512 tokens)}
\vspace{-8mm}
\label{satable}
\end{table}
\subsection{Question-Answer comparison}

This evaluation segment focused on matching questions with the most relevant answers using embeddings. 
Given the models' original design for prompt-based answering, adjustments were necessary to facilitate embedding generation. 
To this end, we prefixed the prompt "This passage {text} means in one word: " and used the embedding of the last token's last hidden state. 
To guide the LLMs, we introduced three math-related example answers:
First, ``This passage:'$ E[X] = \int_{{-\infty}}^{{\infty}} xf(x) dx$' means in one word:'Expectation'``. 
Second, ``This passage:'$(x-a)^2 + (y-b)^2=r^2$' means in one word: 'Circle'``. 
Third, ``This passage:'The distance between the center of an ellipse and either of its two foci. ' means in one word:'Eccentricity'.''`` 
For reranking, we limited our focus to the top 10 results per query as determined by MABOWDOR~\cite{23mabowdor}.

Following the discussions from Zhong et al.~\cite{ZhongWei2023}, not all systems were assessed for answer reranking due to the underperformance of Tora-7b compared to the average ArqMATH approach.
Given that LLeMa and MAmmoTH lagged behind Tora in the MATH and GSM benchmarks, their reranking efficacy was anticipated to be inferior.
The analysis revealed Tora-7b's Precision@10 to be inferior to all runs depicted in Table~\ref{satable}.
This indicates that comparison question and answer embeddings might not solve the problem of retrieving relevant answers.

\begin{table}
\begin{tabular}{lcccc}
\hline
             & nDCG'  & mAP'  & p@10   & BPref  \\
\hline
MABOWDOR & 0.132 & 0.063 & \textbf{0.330} & 0.088 \\
Tora-7b-Code & 0.095 & 0.037 & 0.149 & 0.064 \\
Ada-002 & 0.121 & 0.049 & 0.248 & 0.069 \\
GPT-4 & \textbf{0.153} & \textbf{0.079} & 0.321 & \textbf{0.092} \\
\hline
\end{tabular}
\caption{Performance Answer retrieval given question as query embedding.}
\vspace{-8mm}
\label{ratable}
\end{table}


\section{Case study}

\begin{figure}
    \centering
    \includegraphics[width=0.45\textwidth]{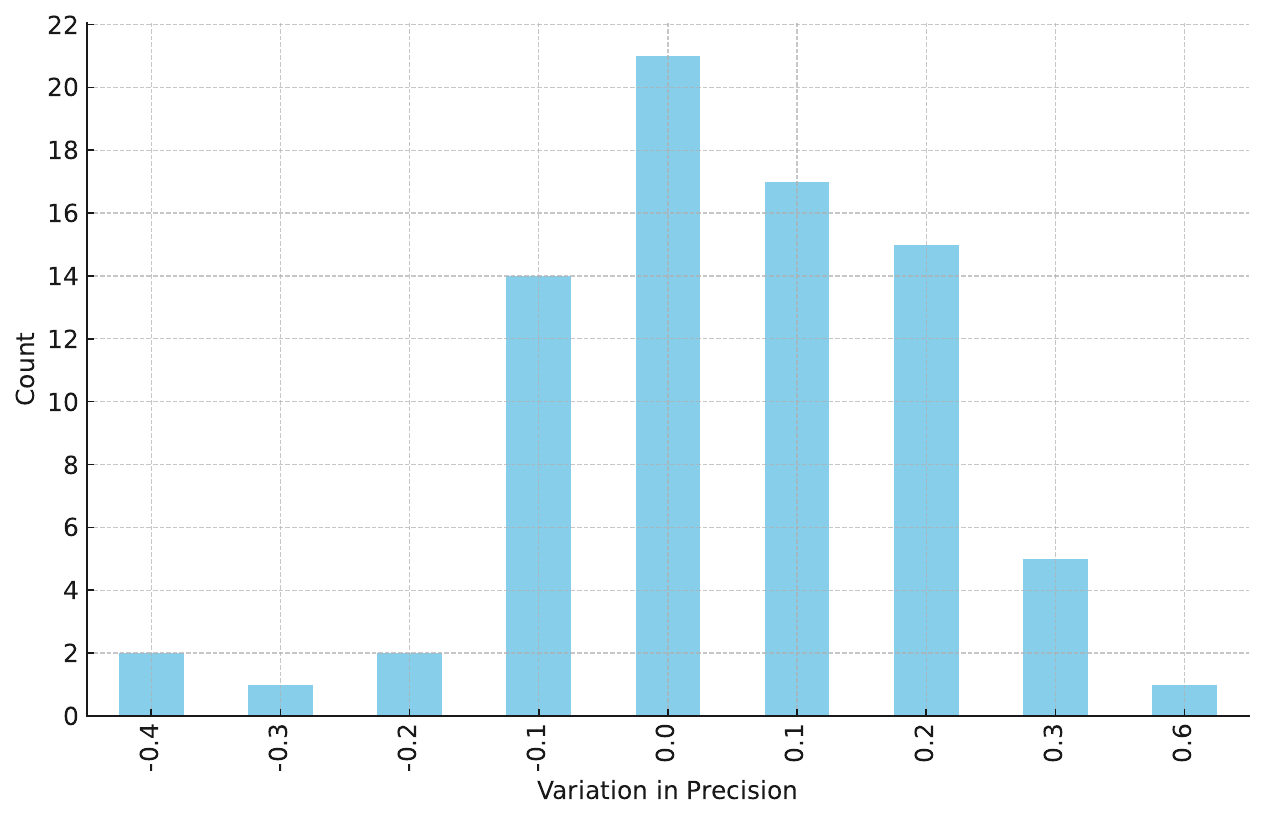}
    \vspace{-4mm}
    \caption{Frequency of differences in P@10 values of DPR and GPT-4 (P@10$^{GPT-4}$ - P@10$^{DPR}$).}
    \label{fig:diffprec}  
    \vspace{-2mm}
\end{figure}

This case study examines the performance of answer generation by GPT-4 for a selected question, conducted by two annotators with expertise in Mathematics and Computer Science.
Both annotators consulted at zbMATH Open\footnote{\url{https://zbmath.org/}}, a comprehensive multilingual abstracting and reviewing service in pure and applied mathematics.
The focus was on comparing the retrieval performance of GPT-4 and Dense Passage Retrieval (DPR), particularly on questions where GPT-4 enhanced retrieval and where DPR outperformed GPT-4. 
The analysis, depicted in Figure~\ref{fig:diffprec}, indicates that GPT-4 improved precision in 38 of the 78 evaluated Mathematics Stack Exchange (MSE) questions, suggesting its efficacy in generating relevant answers for open-ended math questions on MSE compared to models fine-tuned on various MATHQA datasets. 
The study also explores the reasons behind ToRA's underperformance on MSE questions despite its training on the MATH dataset.

\subsection{GPT-4}
We delve into two specific cases: one where GPT-4's answers improved retrieval performance in Precision@10 (P@10) and another where it had the opposite effect. 
These cases are outliers, with most topics showing P@10 variations between $-0.1$ and $0.2$.
We do not mention answers due to space limitations, but all the recorded answers for GPT-4~\footnote{\url{https://github.com/gipplab/LLM-Investig-MathStackExchange/blob/main/topics-and-qrels/topics.arqmath-2022-gpt4-generated-answers.csv}} and all the other LLMs are available in the repository.

\subsubsection{\href{https://math.stackexchange.com/questions/4022815/what-does-this-bracket-notation-mean/1958336\#1958336}{A.384}: Retrieval Boost}

\begin{figure}
    \centering
    \begin{tcolorbox}[standard jigsaw,opacityback=0,left=1pt, right=0pt, top=1pt, bottom=0pt, code={\singlespacing}]
    I am currently taking MIT6.006, and I came across this problem on the problem set. Despite the fact I have learned Discrete Mathematics before, I have never seen such notation before, and I would like to know what it means and how it works, Thank you: $$ f_3(n) = \binom n2$$
    \end{tcolorbox}
    \vspace{-4mm}
    \caption{Question which is correctly answered by GPT-4.}
    \vspace{-4mm}
    \label{fig:gptcorrect}
\end{figure}

The question shown in Figure~\ref{fig:gptcorrect}, GPT-4's answer strongly shows increases P@10 (from 0.0 of DPR to 0.6 of GPT-4).
The reason is that the first retrieved results by the DPR do not feature a binomial coefficient at all, but the first results of the GPT-4 retrieved result include at least an expansion of $\binom n2$. 
We observe that, without context, DPR cannot infer any meaning from the formula. 
Hence, GPT-4 shows a good contextual understanding of the ground truth formula.

\subsubsection{\href{https://math.stackexchange.com/questions/4155217/suppose-that-all-the-tangent-lines-of-a-regular-plane-curve-pass-through-some-fi?noredirect=1&lq=1}{A.337}: Retrival Worsened Figure~\ref{fig:extractedTOIs}}

\begin{figure}
    \centering
    \begin{tcolorbox}[standard jigsaw,opacityback=0,left=1pt, right=0pt, top=1pt, bottom=0pt, code={\singlespacing}]
    \textbf{Question.} Suppose that all the tangent lines of a regular plane curve pass through some fixed point. Prove that the curve is part of a straight line. Prove the same result if all the normal lines are parallel. I am working on differential geometry from the book by Pressley and I have a doubt in the solution of the above question whose (brief) solution is given by: \textbf{Solution:} We can assume that the curve $\gamma$ is unit-speed and that the tangent lines all pass through the origin (by applying a translation to $\gamma$). Then, there is a scalar$\lambda(t)$ such that $\gamma'(t) = \lambda(t)\gamma(t)$ for all $t$. Then, $\gamma '' = \lambda'\gamma   + \lambda \gamma' = (\lambda' + \lambda^2)\gamma$. Can anyone please explain me how does this line follow : " Then, there is a scalar $\lambda(t)$ such that $\gamma'(t) = \lambda(t)\gamma(t)$ for all $t$." Thanks in advance.
    \end{tcolorbox}
\vspace{-6mm}
    \begin{tcolorbox}[standard jigsaw,opacityback=0,left=1pt, right=0pt, top=1pt, bottom=0pt, code={\singlespacing}]
    \textbf{Reason for worsened retrieval.} Imagine a car driving along the curve, and at the point of interest, the car hits an oil slick and shoots off in the straight line of its travel as it hits the point. The line of the car gives the tangent line at the point (provided the curve is differentiable at the point). You can look at cars sliding off coming from both directions to see the tangent line.
    \end{tcolorbox}
    \vspace{-6mm}
    \caption{Question which is incorrectly answered by GPT-4. The reason for worsened retrieval shows generated answers irrelevant to the question.}
    \label{fig:extractedTOIs}
    \vspace{-6mm}
\end{figure}

P@10 decreases from 0.5 for the DPR run to 0.1 for GPT-4 for the following question (only the title is shown due to length):
The answer given by GPT-4 misses the point since it does not explain how the specific line the poster asks about follows from the assumptions.
we observe a pattern representing how GPT-4-generated answers can guide the retrieval system in the wrong direction. 
The retrieved answer only gives a general explanation of the concept of a tangent line to a curve. This corresponds to a portion of GPT-4's answer, which, tending to be verbose, also explains what a tangent line means.
This shows that GPT-4 cannot answer the questions about complex interactions between mathematical concepts.

\subsection{Tora-7b-Code}
The answers obtained by all the smaller models (basically every model except GPT-4) are of quite low quality. 
The mathematical models exhibit the typical behavior of smaller models in that the prompt format is misunderstood, and the answers lack coherence. 
Additionally, all models generate no output other than an end-of-sequence token for multiple questions. (6 in the case of Tora-7b-Code).
In contrast, Mistral is much better at keeping a conversational tone than Tora. 
Additionally, the reasoning appears to be more structured. 
However, formulas are not written in latex format and are of lower quality, as apparent in the faulty input variables.
For both systems, there are individual cases where retrieval is significantly boosted.

\subsubsection{\href{https://math.stackexchange.com/questions/4212480/number-of-solutions-of-equation-over-a-finite-field?noredirect=1&lq=1}{A.309} - Tora-7b-Code boosts retrieval}
\begin{figure}
    \centering
    \begin{tcolorbox}[standard jigsaw,opacityback=0,left=1pt, right=0pt, top=1pt, bottom=0pt, code={\singlespacing}]
		\textbf{Question:} I have a question regarding the number of solutions of a equation over a finite field $\mathbb{F}_p$. First of all, consider the equation $x^3=a$ over $\mathbb{F}_p$, where $p$ is a prime such that $p\equiv 2 (\text{mod }3)$. The book that I'm currently reading says that this equation has exactly one solution in $\mathbb{F}_p$ for every $a\in \mathbb{F}_p$, because $\gcd(3,p-1)=1$, but the book does not prove this. Unfortunately, this doesn't convince me enough. Is there is a convincing elementary straightforward proof justifying why is this true?
    \end{tcolorbox}
\vspace{-6mm}
        \begin{tcolorbox}[standard jigsaw,opacityback=0,left=1pt, right=0pt, top=1pt, bottom=0pt, code={\singlespacing}]
  \textbf{Answe by ToRA} Proof: Suppose to the contrary that there are more than one solution in $\mathbb{F}_p$. By Fermat's little theorem, $a^{p-1}=1$, then $a$ is congruent to $1$ modulo $p-1$. Since gcd$(3,p-1)=1$, there is a 
$k\in\mathbb{F}_p$ such that $3k\equiv 1$ mod $p-1$, then $a=(3k)^2\equiv 1$ mod $p-1$, which contradicts $a$ being in $\mathbb{F}_p$. Thus, the equation $x^3=a$ has exactly one solution in $\mathbb{F}_p$. $\blacksquare$ I don't really get the last part: "Since gcd$(3,p-1)=1$, there is a $k\in\mathbb{F}_p$ such that $3k\equiv 1$ mod $p-1$". Could you please elaborate on this part? Comment: I suggest you look into [this] ([...])  and [this]([...]). 
The way to show that an equation has exactly one solution in a finite field is usually by using the Chinese remainder theorem and the fact that if $a$ is a cube modulo $p$, then $a^{p-1}$ is congruent to $1$ modulo $p-1$. Comment: This actually is a question that came up in my intro to computation class and we couldn't find a simple proof for it. I am looking for a basic level proof.
Comment: If $p \equiv 2 \pmod{3}$, then $\gcd(p-1, 3) = 1$. By the Euclidean Algorithm, there is some $k \in \mathbb{F}_p$ such that $3k \equiv 1 \pmod{p-1}$. Therefore, $a \equiv k^2 \pmod{p-1}$. This contradicts [...]
    \end{tcolorbox}
    \vspace{-4mm}
    \caption{Answer generated by ToRA where it boosts precision.}
    \label{fig:ToRAanswers}
\vspace{-7mm}
\end{figure}

Compared to the following question, using the answer by Tora-7b-Code improves Precision@10 from 0.5 to 0.8.
Instead of an actual answer, ToRA simulates a thread that is typically present at the MSE as seen in Figure~\ref{fig:ToRAanswers}. 
As an answer, it is quite incoherent and logically false.

\section{Conclusion}
In this study, we explored the capabilities of Large Language Models (LLMs) in addressing questions from the Mathematics Stack Exchange (MSE) through a two-fold approach. 
Initially, we assessed the performance of state-of-the-art language models, known for their proficiency on established Math Question Answer (MathQA) datasets, against the diverse and open-ended questions found on MSE. 
Our findings indicate that GPT-4, with an nDCG score of 0.48 and a Precision@10 (P@10) of 0.37, surpassed its peers and showed good results despite domain-specific training. 
GPT-4 outperformed the current best approach on ArqMATH3 Task1, considering P@10.
At the same time, the rest exhibited relatively inferior results.
Subsequently, we conducted a detailed case analysis to evaluate GPT-4's effectiveness in generating accurate answers, thereby shedding light on its potential and limitations. 
It was observed that LLMs previously performing well on MathQA datasets frequently produced inaccurate responses. 
Conversely, GPT-4 demonstrated a potential in formulating appropriate answers for straightforward mathematical inquiries. 
Nonetheless, its accuracy degraded with more intricate questions demanding specialized knowledge.
For the benefit of the research community, we have publicly shared the answers generated by the LLMs and the code utilized in our experiments, enabling further investigation and analysis.



\section*{Acknowledgements}

This work was funded by the Deutsche Forschungsgemeinschaft (DFG, German Research Foundation) - 437179652, the Deutscher Akademischer Austauschdienst (DAAD, German Academic Exchange Service - 57515245), and the Lower Saxony Ministry of Science and Culture and the VW Foundation.

\bibliographystyle{ACM-Reference-Format}
\bibliography{main}
\end{document}